\documentclass[10pt,twocolumn,letterpaper]{article}

\usepackage{iccv}
\usepackage{times}
\usepackage{epsfig}
\usepackage{graphicx}
\usepackage{amsmath}
\usepackage{amssymb}
\usepackage{multirow}
\usepackage{booktabs}  
\usepackage{threeparttable}  
\usepackage{color}
\usepackage{array}
\usepackage{makecell}
\usepackage{stfloats}

\usepackage[pagebackref=true,breaklinks=true,letterpaper=true,colorlinks,bookmarks=false]{hyperref}

\iccvfinalcopy 

\def\iccvPaperID{9839} 

\ificcvfinal\fi

\begin{document}

\title{Target Transformed Regression for Accurate Tracking}


\author{Yutao Cui \and
Cheng Jiang \and
Limin Wang\thanks{Corresponding Author.} \and
Gangshan Wu \and
State Key Laboratory for Novel Software Technology, Nanjing University, China
}


\maketitle
\ificcvfinal\thispagestyle{empty}\fi

\begin{abstract}

Accurate tracking is still a challenging task due to appearance variations, pose and view changes, and geometric deformations of target in videos. Recent anchor-free trackers provide an efficient regression mechanism but fail to produce precise bounding box estimation. To address these issues, this paper repurposes a Transformer-alike regression branch, termed as {\em Target Transformed Regression} (TREG), for accurate anchor-free tracking. The core to our TREG is to model pair-wise relation between elements in target template and search region, and use the resulted target enhanced visual representation for accurate bounding box regression. This target contextualized representation is able to enhance the target relevant information to help precisely locate the box boundaries, and deal with the object deformation to some extent due to its local and dense matching mechanism. In addition, we devise a simple online template update mechanism to select reliable templates, increasing the robustness for appearance variations and geometric deformations of target in time.
Experimental results on visual tracking benchmarks including VOT2018, VOT2019, OTB100, GOT10k, NFS,
UAV123, LaSOT and TrackingNet demonstrate that TREG obtains the state-of-the-art performance, achieving a success rate of 0.640 on LaSOT, while running at around 30 FPS. The code and models will be made available at
\href{https://github.com/MCG-NJU/TREG}{https://github.com/MCG-NJU/TREG}.

\end{abstract}

\section{Introduction}

Visual object tracking~\cite{siamfc,siamrpn,atom,dimp} is an important yet challenging task in computer vision with a wide range of applications, such as robotics, surveillance~\cite{introduction2}, and human-computer interaction~\cite{introduction1}. It aims to estimate the state of an arbitrary object in video frames, given the target bounding box in an initial frame. Although much progress~\cite{siamfc,siamrpn,atom,dimp} had been made in recent years, accurate tracking still remains challenging, due to the fact that the target might be with deformation, pose and viewpoint changes, and even occluded by other objects. In general, tracking a single object can be decomposed into sub-tasks of classification and regression, which is to localize the target roughly and regress the precise bounding box, respectively.

\begin{figure}[t]
\centering
\includegraphics[width=8.3cm]{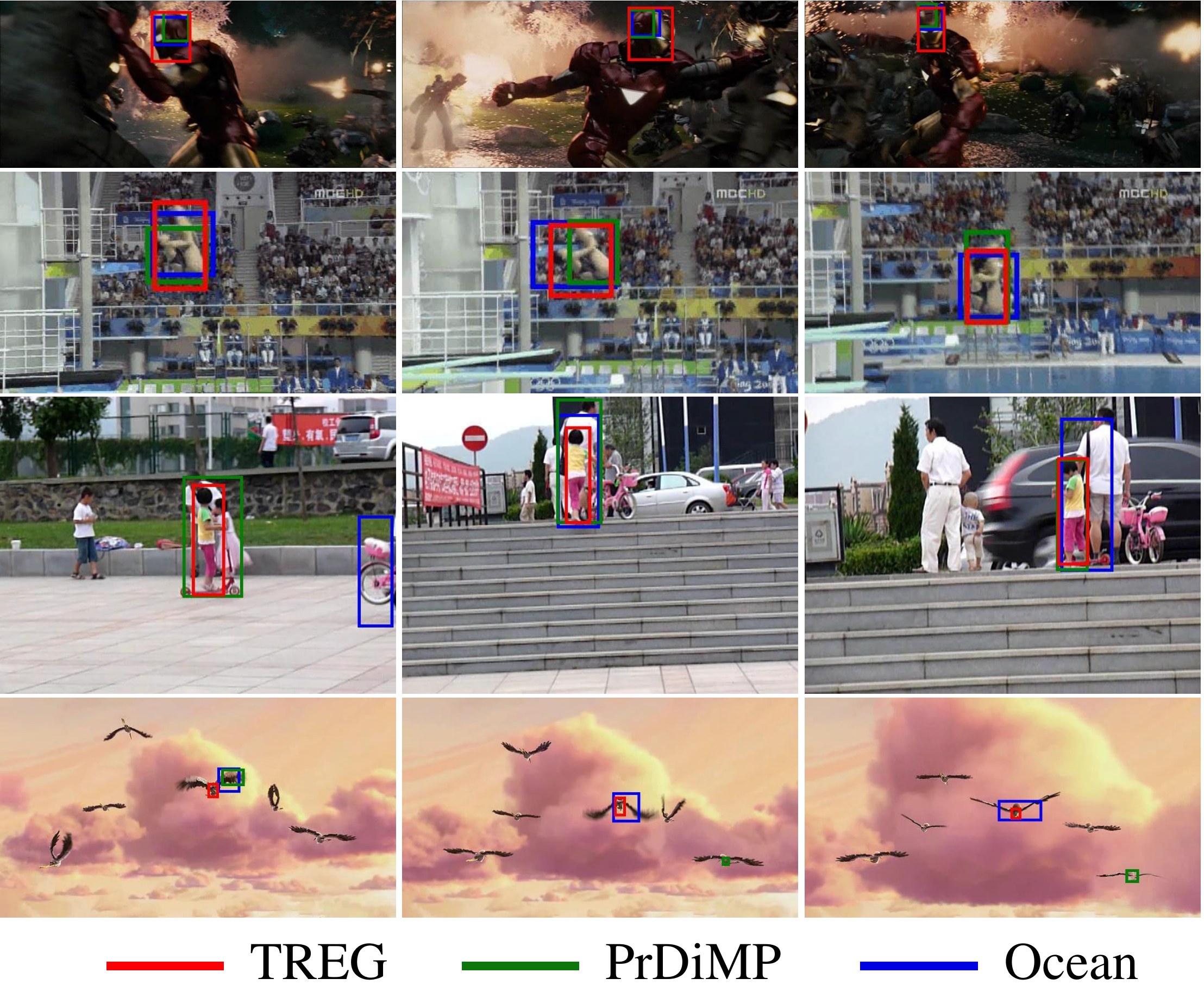}
\caption{A comparison of our approach with state-of-the-art tracker. Observed from the visualization results, our TREG produces more precise bounding boxes than the state-of-the-art trackers PrDiMP~\cite{prdimp} and other anchor-free trackers, such as Ocean~\cite{ocean}, when encountering circumstances of deformation, scale changes and fast movement.
}
\label{fig:1}
\end{figure}

In order to build an accurate tracker, regression branch design is of great importance as it is responsible for generating the precise bounding box. The previous works on box estimation can be roughly grouped into two types: (1) indirect bounding box estimation and (2) direct bounding box regression. For the first type, early approaches~\cite{siamfc} simply use a multi-scale searching strategy based on the classification branch. Then, ATOM~\cite{atom} method presents a specialized IoU prediction network to select and refines the final object box. For the second type, the seminal Siamese trackers~\cite{siamrpn,dasiamrpn,siamrpnPlus} resort to anchor-based mechanism to direct regress the bounding size based on the predefined anchors. More recently, some anchor-free trackers~\cite{ocean,siamfc++,siamban,siamcar} are more popular owning to its simplicity in design and superior performance, by directly regressing the box size. However, unlike image object detection, these anchor-free trackers are not sufficiently accurate and robust for object tracking, due to the essential illness of tracking problem that we expect a tracker to be trained with a one-shot supervision yet to generalize well to unseen deformations and variations.

In this paper, our objective is to design a more accurate anchor-free tracker by proposing a customized regression branch to effectively handle object deformation, pose and view changes. Based on the above analysis, we argue that the core problem of accurate anchor-free tracking is {\em how to integrate the target information to regression branch to retain its precise boundary information and handle its variations in time}. The existing anchor-free trackers~\cite{ocean,siamfc++,siamban,siamcar} simply utilize a depth-wise correlation representation to fuse target information or a target guided attention module to modulate the visual representations. Although these techniques provide feasible solutions for target information integration, they may contain insufficient target information for precise regression, and lack flexibility to deal with object variations. Accordingly, we figure out two important factors that need to be considered specifically. First, in order to generate precise object boundaries, we need to keep sufficient target information in regression branch. Second, for handling appearance variations and object deformation, regression branch is expected to be flexible with deformation and adaptive over time.


Following the above analysis, we devise an anchor-free and target-guided regression branch with a transformer-alike design, termed as Target Transformed Regression (TREG), partially inspired by the success of Transformer~\cite{transformer} in context modeling. Basically, we utilize the cross-attention mechanism in transformer to explicitly model all pairwise interactions between elements of target template and search areas, making these enhanced representations particularly suitable for precise boundary offset regression. Specifically, the feature cells in target template are encoded as \textit{key} and \textit{value} at first. Then, for each position in search area, we enhance its visual representation by querying its feature over all the pairs of \textit{key-val} in the target template. This retrieved target-aware representation is able to enhance the target-relevant information and deal with object deformation to some extent, thanks to the local and dense matching between all elements of target template and search regions. To further improve the effectiveness of our TREG, we establish an online target template queue to maintain the variations of the tracked object, and design a confidence based update strategy to adaptively select reliable target templates. Finally, we place a feed-forward network on top of the target transformed representation to perform object boundary offset regression. 

Combined with the existing online classification branch from DiMP~\cite{dimp}, we develop a principled anchor-free tracking framework. We perform comprehensive experiments on eight benchmarks~\cite{vot2018,vot2019,got10k,otb,trackingnet,lasot,uav123,nfs} to demonstrate the superior performance of our TREG to previous state-of-the-art methods. The contributions of this work are three-fold:
\begin{itemize}
\item We propose an accurate anchor-free tracker by specifically devising a target transformed regression branch (TREG). The advantage of modeling pair-wise relation between elements in target template and search area enables our TREG to keep precise boundary information and effectively deal with objects variations.
\item We present a simple online target update mechanism by establishing a confidence based template queue, which enables the tracker to be flexible to deal with appearance variations and geometric deformations of object over time. 
\item Our TREG outperforms the popular state-of-the-art real-time trackers on eight benchmark datasets including VOT2018~\cite{vot2018}, VOT2019~\cite{vot2019}, LaSOT~\cite{lasot}, TrackingNet~\cite{trackingnet}, OTB~\cite{otb}, GOT10k~\cite{got10k}, UAV123~\cite{uav123} and NFS~\cite{nfs}, especially achieving a success rate of 0.640 on LaSOT, while running at a real-time speed of around 30 FPS.
\end{itemize}

\section{Related Work}
In this section, we briefly introduce recent trackers from the aspect of target regression. Besides, we discuss the transformer mechanism used in tracking.

\begin{figure*}
\begin{center}
\includegraphics[width=13.5cm]{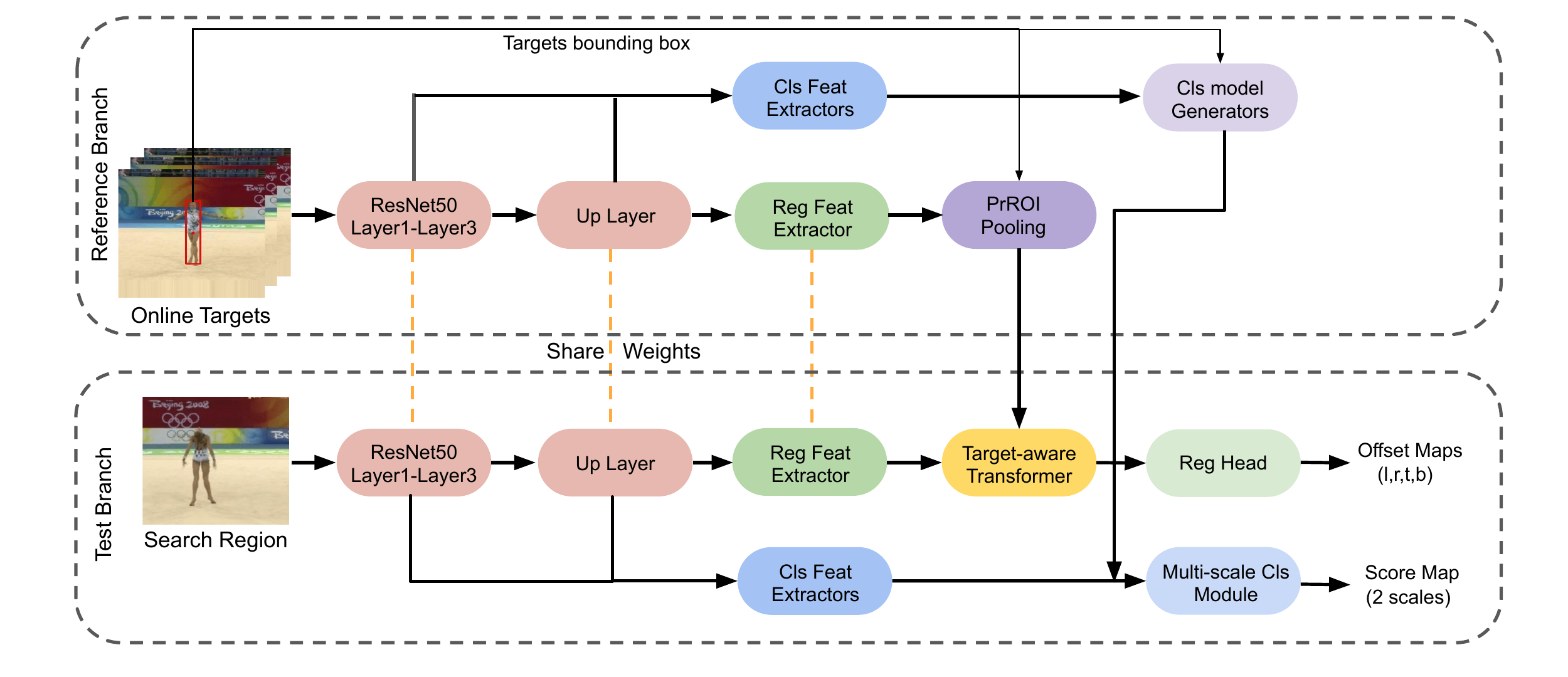}
\end{center}
   \caption{{\bf TREG Framework}.
We present an end-to-end anchor-free tracker, termed as TREG, for accurate tracking, which is composed of a backbone to extract common features, classification feature extractors and a regression feature extractor to extract task-specific features, multi-scale classification module and target-aware transformer based regression module to localize the target center and estimate the precise target bounding box respectively.
Similarly to FCOT~\cite{fcot}, We employ a multi-scale online classification component, where the discriminative model generator is proposed in DiMP~\cite{dimp}, to localize an accurate target center. 
For regression, a simple online updating target-aware transformer is proposed to yield a robust and precise regressor. 
Detailed structure of the online target-aware transformer can be found at Figure.~\ref{fig:transformer}. }
\label{fig:architecture}
\vspace{-0.2cm}
\end{figure*}

\paragraph{Target regression for tracking} 
Target regression is employed in estimating the precise target state. Previous works can be roughly grouped into two types, indirect target estimation and direct bounding box regression. For the former one,
some CF-based trackers~\cite{kcf,eco,dcf_,mdnet} and SiamFC~\cite{siamfc} employed brutal multi-scale test to estimate the target scale roughly. ATOM~\cite{atom} and DiMP~\cite{dimp} employ a specialized IoU prediction network to select and refine the final object box. 
For the latter one,
RPN-based trackers~\cite{siamrpn,siamrpnPlus,dasiamrpn,siamDW,StructSiam,siammask} regress the location shift and size difference between pre-defined anchor boxes and target location. SATIN~\cite{SATIN} and CGACD~\cite{CGACD} detect corners using cross-correlation operation and correlation-guided attention operation respectively. SiamFC++~\cite{siamfc++} directly regresses the offset to box corners. Anchor-free trackers~\cite{siamban,siamcar,ocean,fcot} are more and more popular owning to its simplicity in design and superior performance. 
However, to use target information, the existing anchor-free trackers employ a target-guided attention or a depth-wise correlation to modulate the search frame representation, which may contain insufficient target information for precise regression, and lack flexibility to deal with object variations. We devise a target transformed regression branch to tackle these issues and acquire superior performance.

\paragraph{Transformer mechanism in tracking}
Generalized transformer mechanism~\cite{transformer} is a group of neural network layers that aggregates information from the entire input sequence. It introduces attention layers, which scan through each element of a sequence and update it by aggregating information from the whole sequence. 
In the area of visual tracking, CSR-DCF~\cite{Lukezic_2017_CVPR} constructs an object spatial attention map to constrain correlation filter learning and calculates the channel reliability values of weighted sum correlation
response maps. Then RASNet~\cite{rasnet} introduces spatial and channel-wise attention to a Siamese network. CGCAD~\cite{CGACD} further proposes a correlation-guided attention for corner detection. However, the pixel-wise correlation-guided spatial attention overlooks the fact that there are some background parts in the target, which may result in high attention weights outside the object region.
Consequently, they adopt a complex two-stage structure for estimating one RoI and then detecting the corners, so as to alleviate the issue. 
Compared with them, our target-aware transformer retains sufficient target information to enhance the regression representation and can be easily deployed for online regression thanks to its simple structure.

\section{Proposed Method}

We develop a target transformed regression branch with online updating mechanism for accurate anchor-free tracking. The online target transformed regression component is basically devised complying with the following guidelines:
(\romannumeral1) a fully target integration module to yield high-quality visual representation to retain sufficient target information for precise object boundary generation, (\romannumeral2) pixel-wise context modeling to enhance the target-relevant features and cope with object deformation, and (\romannumeral3) an efficient online mechanism so as to handle appearance variations in a consecutive sequence.

We take inspirations from the transformer in context modeling and its variants~\cite{transformer,nonlocal} in computer vision, and devise a transformer-alike structure for anchor-free regression. Given a query element and a set of key elements, a transformer function adaptively aggregates the key contents to transform the query feature based on attentive weights that measure the compatibility of \textit{query-key} pairs.
Following this technical perspective, a target-aware transformer is designed for incorporating target appearance into our regression features. The feature cells of target template extracted by a backbone and an ROI pooling layer is encoded as \textit{key} and \textit{value} elements. Then for each position in search region, we enhance its visual representation by querying its feature over all the pairs of \textit{key-value}. Then the retrieved representations are fused to original feature so that we can obtain rich target-aware information for regression.
Through the local and dense matching between all elements of target template and search regions, we can not only preserve sufficient target information to produce high-quality representation for regression, but also be able to handle with object deformation to some extent. 
Additionally, we establish an online target template queue to maintain the variations of the tracked object, composed of three static targets augmented by the given template and reliable online targets adaptively selected based on the classification confidence. 
Finally, a feed-forward network is placed on top of the target transformed representation to perform object boundary offset regression directly.
By these designs, we construct a simple yet accurate target transformed regression branch.

Based on our proposed target transformed regression branch, we build an efficient and principled anchor-free tracker (TREG), by combining an online classification branch. The framework of the proposed tracker is shown in Figure.~\ref{fig:architecture}. The whole framework is composed of a common backbone to extract features for both classification and regression branches, a classification branch to localize the target center, and a regression branch to estimate precise target state.
Similar to FCOT~\cite{fcot}, we employ a multi-scale online classification component, where the discriminative model generator is proposed in DiMP~\cite{dimp}, to estimate a robust target center location.
Given the target center location and template in online targets queue, the target-aware transformer produces the enhanced visual features, which are used to regress offset to the object boundary directly.
We describe the target transformed regression branch and the framework in details in the next subsections. 

\begin{figure*}[t]
\begin{center}
\includegraphics[width=13.5cm]{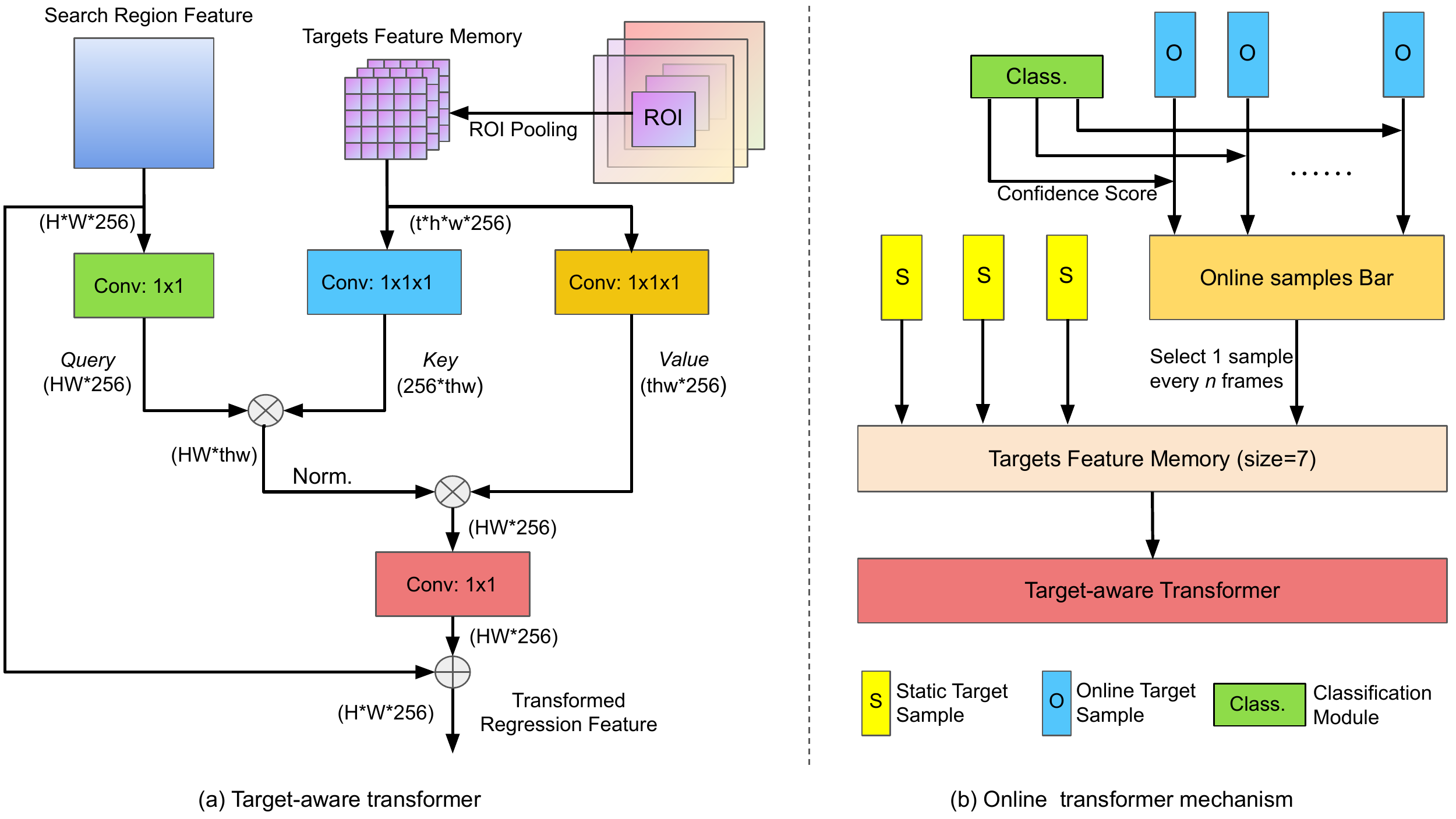}
\end{center}
   \caption{{\bf Online Target-aware Transformer for Regression}. 
   (a) Target-aware transformer takes search region feature and pooled targets feature as input, producing transformed regression feature. Two 3D $Conv$ with kernel size of $1 \times 1 \times 1$ is performed to targets feature to generate $key$ and $value$ respectively. Similarly, a 2D $Conv$ with kernel size of $1 \times 1$ is performed to search region feature to yield $query$ element. "$\bigotimes$" denotes matrix multiplication, and "$\bigoplus$" denotes element-wise sum. $Norm.$ denotes scaling by $1/(t \times h \times w)$. (b) Online template update mechanism is implemented by maintaining a targets template queue with size of 7, which is composed of 3 static targets feature and 4 online updating ones. The static targets is acquired by performing an augmentation to the given target. The online targets memory is updated every $n$ frames. An online samples bar is to reserve $n$ targets feature with classification scores. Then the proper online target is selected by maximize the confidence scores from the samples bar.
   }
\label{fig:transformer}
\vspace{-0.2cm}
\end{figure*}

\subsection{Target transformed regression}

In general, anchor-free regression is solely based on the appearance representation to predict the geometric offsets of box boundary with respect to object center by using a feed-forward network. In order to bridge the gap between appearance information and geometric structure, we hope to incorporate more detailed structure information (e.g., object boundary) into our visual representation, thus relieving the difficulty of geometric structure prediction. As for the specific anchor-free object tracking, we aim to leverage more detailed target information to enhance the object relevant areas while suppress the distractor from background. However, the depth-wise correlation~\cite{siamban,siamfc++,ocean,siamcar} yields similarity map with inconspicuous boundary of the target facing with object deformation since the whole target is served as a correlation filter. Meanwhile, the pixel-wise correlation-based attention~\cite{CGACD} imposes other challenge due to retaining insufficient target information. The region outside the object may be assigned with a high attention weight when there exists background parts in target.
To overcome these shortcomings, we choose to better exploit the all pair-wise correlation between the elements in target and search area, and aim to leverage this rich and high-order features to keep sufficient target-related information, and as well handle the object variation issues during tracking.

\paragraph{Target-aware transformer.}
As outlined in the above analysis, our target transformed regression is inspired and re-purposed from the Transformer architecture~\cite{transformer}.
Basically, we use the search area as a query to enhance its representation with target information.
Concretely, the target hidden representation is encoded as \textit{key} and \textit{value} elements in a pixel-wise way, providing weighted aggregation response for a \textit{query} which is a position of search region feature. 
We define the target-aware feature transformation as:
\begin{equation}
\begin{split}
   y_{i} = W(\frac{1}{N} \sum_{k} \sum_{j\in\Omega_{k}} A(t_{j}, x_{i})\odot \omega(t_{j})) + x_i.
\end{split}
\end{equation}
Here, $i$ indexes a position in search region with representation $x_i$. $j$ is the index that enumerates all possible positions in target template with feature representation $t_j$. $k$ indexes the template in our target queue and $\Omega_{k}$ specifies the feature cells in target template for query. The function $\omega$ computes
a representation of the target signal at the position $j$ which is served as \textit{value} element. A pairwise function $A(t_{j}, x_{i})$ depicts the relations between $t_j$ and $x_i$. The implementation form of $A(t_{j}, x_{i})$ is as follows:
\begin{equation}
\begin{split}
  A(t_{j}, x_{i}) = \theta(x_i)^{T}\phi(t_j),
\end{split}
\end{equation}
where $\theta$ function encodes $x_i$ as a \textit{query} element and $\phi$ function encodes $t_j$ as a \textit{key} element. Target-aware information aggregation is with a weighted sum.
Then the weighted sum is scaled by $1/N$ to perform normalization. $N$ is the total number of elements in our target queue, which is $t \times h \times w$ and $t$ is the number of template and $h$ and $w$ is template size. Particularly, the normalization factor $1/(t \times h \times w)$ can not be substituted with Softmax function as in \cite{nonlocal,transformer}. The reason lies in that some positions in background and distractors of the search region are expected to have low dependency with target, while Softmax function will amplify this noise influence as the sum of attention weights between the \textit{query} and all the \textit{key}s is always 1.
Besides, $W$ represents feature transform to make the queried feature to be the same shape with $x_i$. $y_i$ is the target transformed representation with a simple average of original features and retrieved representation.

As in Figure.~\ref{fig:transformer}, target-aware transformer takes search region feature and ROI-pooled target feature as input, where the spatial size of search region feature is $88 \times 88$ and the target size is $5 \times 5$ in this work. The functions $\phi$ and $\omega$ are implemented with a 3D convolution layer with kernel size of $1 \times 1 \times 1$ respectively, since an online 3D targets memory is maintained as described in ~\ref{online_trans_sect}. Similarly, the functions $\theta$ and $W$ are 2D convolution layers with kernel size of $1 \times 1$.
A target-aware transformer operation is a flexible building block and can be easily inserted to the current anchor-free trackers. There are only 4 convolution layers with kernel size of 1 to be trained offline in the block. Furthermore, we can keep a target queue of variable sizes, which can be easily deployed for online updating.

\paragraph{Discussion with other anchor-free regression.}

To better expound the insight of our target-aware transformer, we compare with the regression component of previous anchor-free trackers. In general, there are the two major methods for integrating target information to search region feature in anchor-free regression. 
First, most of them ~\cite{siamfc++,siamcar,siamban,ocean} employ depth-wise correlation to produce similarity map based on the whole template. Since the correlation filter simply possesses global information of the target, it is difficult to reflect the object boundary precisely when encountering deformation or similar intricate circumstances. 
Secondly, CGACD~\cite{CGACD} brings in a complex correlation-guided attention for detecting the corner which imposes other limitations. Pixel-wise correlation is the primary operation to generate target-aware spatial attention. However, it overlooks the fact that there are some background parts in the target, which may result in high attention weights outside the object region. Consequently, they adopt a complex two-stage structure for estimating one RoI expected to contain the target and then detecting the corners, so as to alleviate the issue.

In contrast, our target-aware transformer explicitly models all pair-wise interactions between elements of target template and search area, and uses the retrieved feature representation to enhance the original features. This unique design makes our target transformed representation to attain more detailed target-relevant information than previous methods, and well handle the appearance variation and deformation of object due to our local and dense matching mechanism. The exploration study demonstrates that our target transformed regression is able to yield more accurate tracking results, as discussed in Section~\ref{analysis}.

\subsection{Online template update} \label{online_trans_sect}

In order to cope with the target variations in a consecutive sequence, we present an online template update mechanism for regression as visualized in Figure.~\ref{fig:transformer}(b). An inherent issue that comes with the online scheme, is that the tracked objects may be imprecise. The tracker is thus confused and tends to drift. It is crucial for accurate regression to make a trade-off between using online targets and static targets. Therefore, an online targets template queue comprising 3 static targets and 4 online targets is maintained. The static targets are acquired by performing a data augmentation to the given template. The online targets memory is updated every $n$ frames, where $n$ is the updating interval. Since the tracked targets are unstable, we design a confidence based updating strategy to adaptively select reliable target templates.  When the object is predicted with the maximum confidence among the online targets in the samples bar, its template will be added into target queue.
Experimental results demonstrate the effectiveness of online template update mechanism in Section~\ref{analysis}.

\subsection{Implementation details}
\paragraph{Framework.}
Following the architecture of ~\cite{dimp}, we employ a ResNet-50~\cite{resnet50} as the backbone to extract common feature. Then an $UP$ layer, composed of 2 convolutional layers and 2 up-sampling layers are used to generate high-resolution feature. 
Then the Classification head and the regression head extract task-specific features to cope with classification and regression tasks separately. The classification head and the regression head are composed of a convolutional layer and 2 deformable convolutional layers~\cite{dai2017deformable}.

We employ the multi-scale classification branch similar with~\cite{fcot}, learning discriminative models proposed in~\cite{dimp}. At first, Low-resolution score map and high-resolution score map are generated by separate online classifiers. Then these two maps are fused to predict a robust and precise target center. During training, the classification objective is a Gaussian function map centered at the ground-truth target center. 

The regression branch is composed of the presented target-aware transformer and a feed-forward network containing 2 convolutional layers and a deformable convolutional layer to estimate the offset from a point to the target boundary. During the offline training process, we perform regression for positions in the vicinity of the target center, which is an area with a radius of 2 in this work.

\paragraph{Offline training.}
Similar with some popular trackers~\cite{dimp,prdimp,atom,fcot}, we use the training splits of LaSOT~\cite{lasot}, TrackingNet~\cite{trackingnet}, GOT-10k~\cite{got10k} and COCO~\cite{coco}. We train the model for 50 epochs by sampling 40000 videos per epoch, giving a total training time of 100 hours on 8 Nvidia Tesla V100 GPUs. The whole network is trained end-to-end with a mini-batch size of 80.
We use ADAM~\cite{adam} with learning rate decay of 0.2 at the epoch of 25 and 35. The classification training loss settings are the same with DiMP~\cite{dimp}. For regression, we use IoU loss and the loss weight is set to 1.

\paragraph{Online tracking.} We perform data augmentation to the first frame with translation, rotation,and blurring, yielding a total of 15 initial online training samples for online classification. And we select 3 samples as the static targets for the online target template queue. During the tracking process, we employ the designed confidence based updating strategy to select reliable samples. We ensure a maximum queue size of 7 by discarding the oldest sample. Online classification is performed as in~\cite{fcot}.

\section{Experiments}

Our tracking approach is implemented in Python based on PyTorch. For inference, we test our tracker on a single Nvidia RTX 2080Ti GPU, achieving a tracking speed of around 30 FPS. The code for training and inference will be made public.

\subsection{Exploration study}\label{analysis}
Here, we describe the experiments to analyze the impact of the target transformed regression and the online template updating mechanism for the transfromer proposed in this work. The experiments are performed on the combined UAV123~\cite{uav123} and NFS~\cite{nfs} datasets, as in ATOM~\cite{atom}. There are a total of 233 challenging videos. We report two metrics, area-under-the-curve(AUC) score and precision score (Precision). For fair comparison, all the following experiments are performed under identical training setting.

\begin{figure}[tp]
\centering
\includegraphics[width=8.3cm]{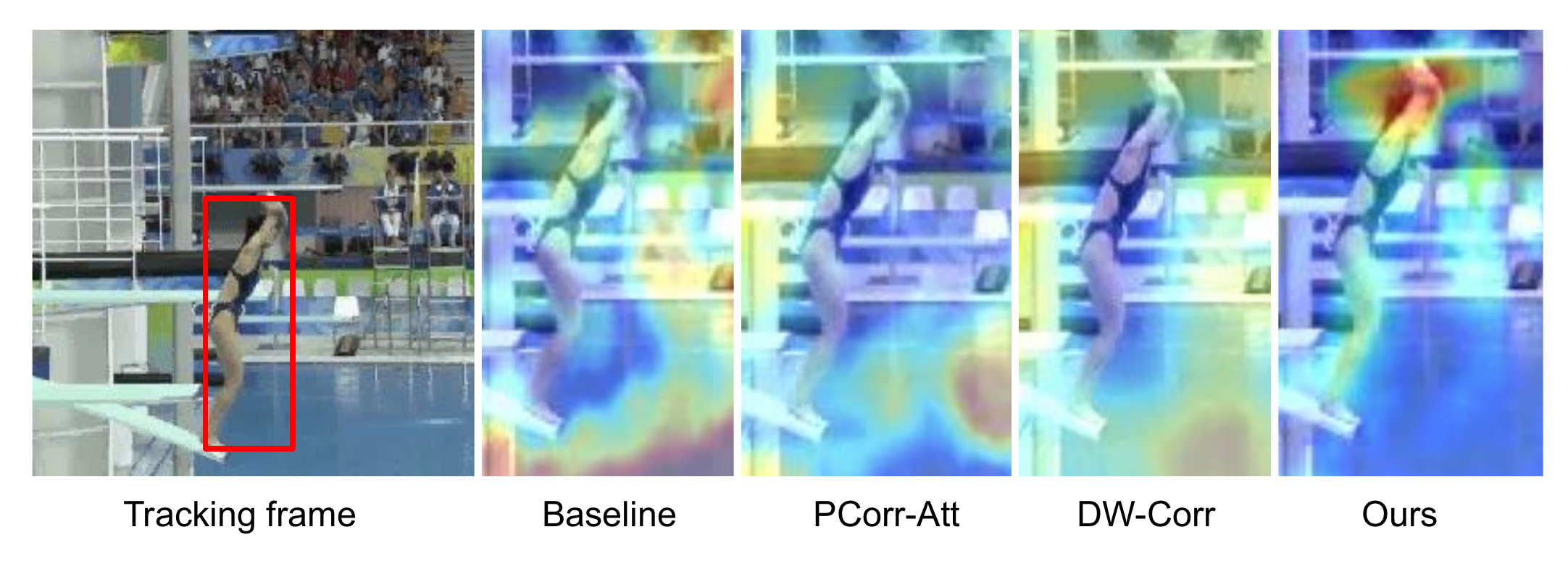}
\caption{Visualization of regression feature for different methods. Red box denotes the ground-truth bounding box.
}
\label{fig:reg_feat_vis}
\vspace{-0.2cm}
\end{figure}

\begin{table}[tp]
\begin{center}
\fontsize{8}{9}\selectfont  
\setlength{\tabcolsep}{1mm}{
\begin{tabular}{cccccc}
\toprule  
    \centering
                &Baseline&DW-Corr&PCorr-Att & TAT-Cls & Ours\cr
    \midrule

    Precision(\%)   &80.5&82.3&82.0 & 82.8 &{\bf 84.0}\cr
    AUC(\%)     &61.6&64.6&62.6 & 64.7 &{\bf 66.0}\cr

    \bottomrule  
\end{tabular}}
\end{center}
\vspace{-0.1cm}
\caption{Analysis of target aware transformer.} 
\label{tab:ab1}
\vspace{-0.4cm}
\end{table}

  





\paragraph{Target-aware transformer for regression.} Our primary contribution is the proposed target transformed regression branch. To evaluate its effectiveness, we take comparison with the following analogous methods for anchor-free tracking. For a fair comparison, We only substitute the target-aware transformer with the following modules, while the other components including classification branch and feed-forward network are the same. And our TREG only utilize the given target as template without the online samples. \textbf{Baseline:} we compare with a baseline approach that removes the target-aware transformer. As shown in Table.~\ref{tab:ab1}, the performance drops by 3.5\% in terms of Precision and 4.4\% in terms of AUC, which indicates that incorporating the target information into regression representation is of vital importance. \textbf{DW-Corr:} then we take comparison with depth-wise correlation based regression method as being commonly used in~\cite{ocean,siamban,siamcar,siamfc++}. The performance has a obvious drop compared with ours, since it can not handle with target deformation for accurate regression. \textbf{PCorr-Att:} we also compare with the pixel-wise correlation-guided spatial attention module proposed  in~\cite{CGACD}, since it has high relevance with the target-aware transformer. It generates inferior performance than ours, which demonstrates that retaining sufficient target information for regression representation is essential for eliminating the interference of the background or distractors. In Figure.~\ref{fig:reg_feat_vis}, we further provide \textbf{\textit{intuitive visualization samples for regression feature}} to illustrate that the proposed method represents the boundary of targets better than other methods. It can be seen that the boundary of the diver including head and foot are get augmented through our target-aware transformer although the object changes in the sequence. 

\begin{figure}[tp]
\centering
\includegraphics[width=8.3cm]{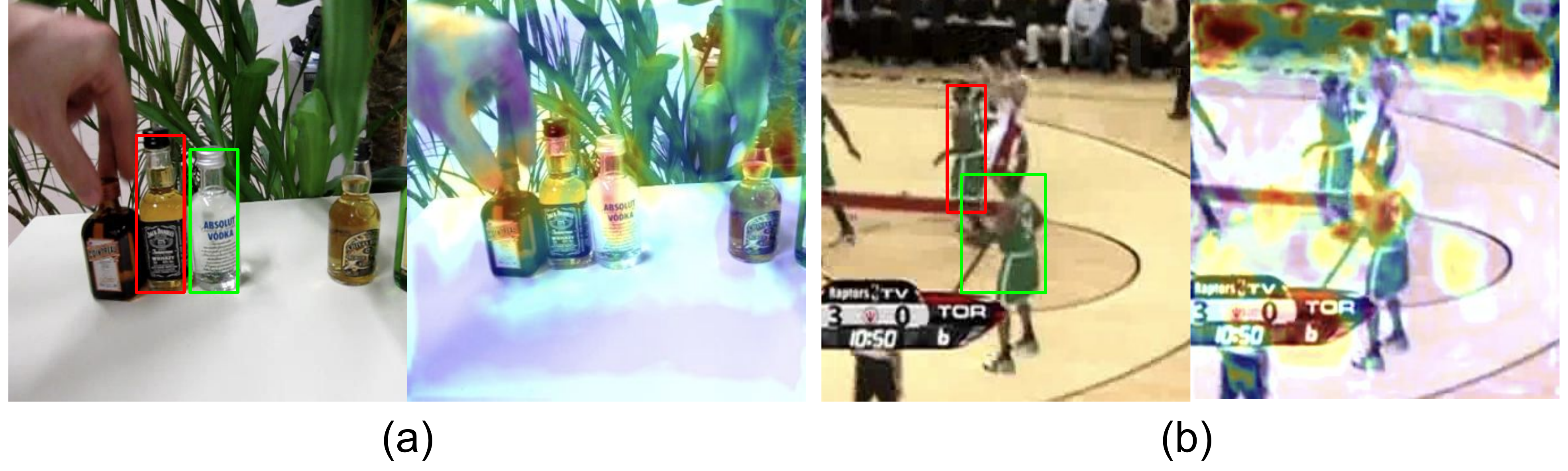}
\vspace{-0.4cm}
\caption{Visualization of classification feature  when applying our target-aware transformer to classification branch. Red box denotes the ground-truth bounding box and the green one denotes the tracked negative object.
}
\label{fig:cls_feat_vis}
\end{figure}

\paragraph{Target-aware transformer for classification.}
We further analyze the appliance of target-aware transformer for classification branch, only substituting the high-resolution classification module with the proposed transformer-alike structure. We can see from Table.~\ref{tab:ab1} that the performance of \textbf{TAT-Cls} is not so good as ours TREG. Specifically, the pixel-to-pixel matching method is not suitable for discriminating similar objects since it tends to overlook the overall information of targets, which can be derived from Figure.~\ref{fig:cls_feat_vis}. As a consequence, target-aware transformer is not employed for classification in our design.


\begin{figure}[t]
\vspace{-0.2cm}
\centering
\includegraphics[width=8.3cm]{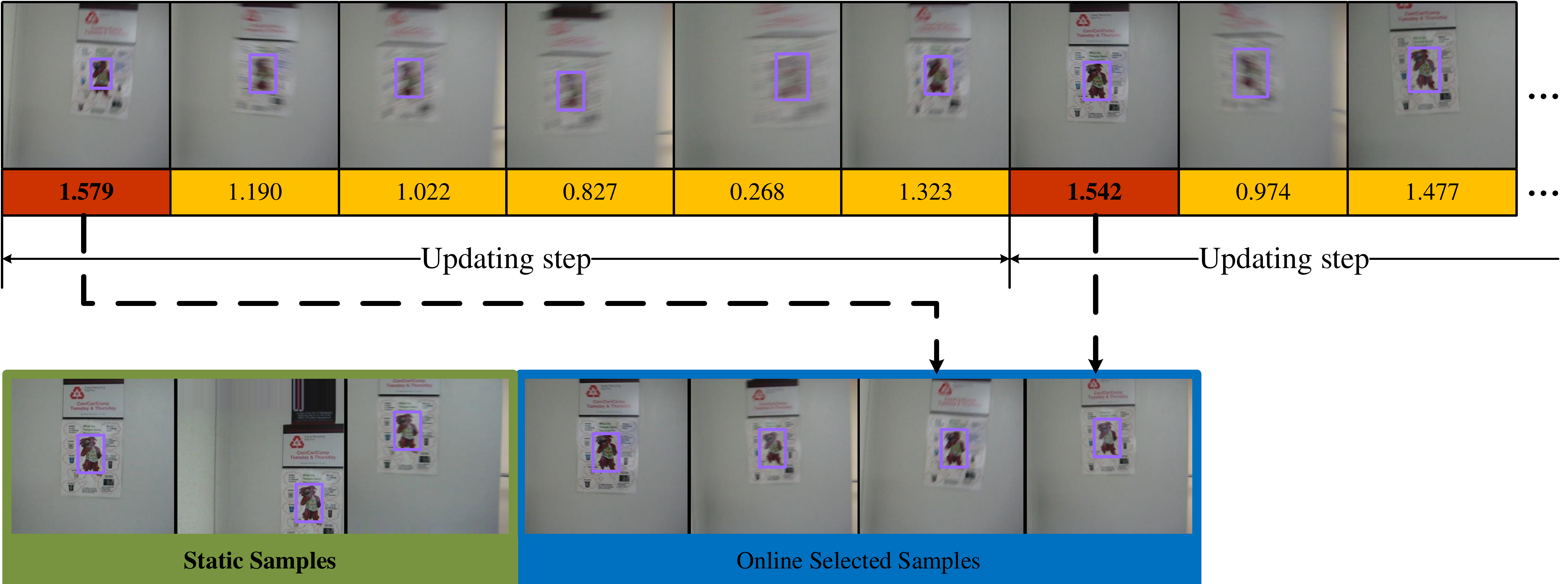}
\caption{Visualization of online targets selection. }
\label{fig:online_queue}
\end{figure}

\setlength{\tabcolsep}{8pt}
\begin{table}[pt]
\begin{center}
\fontsize{8}{9}\selectfont  
\setlength{\tabcolsep}{1mm}{
\begin{tabular}{cccc}
\toprule  
    &Model&AUC&Precision\cr  
    \midrule
    \multirow{3}{*}{Static}&
    queue size=1&66.0&84.0\cr
    &queue size=3&66.3&84.7\cr
    &queue size=7&65.8&83.8\cr
    \midrule

    \multirow{2}{*}{Online}&update with fixed interval&65.6&83.6\cr
    &confidence based update&{\bf 67.2}&{\bf 85.9}\cr
    \bottomrule   
\end{tabular}}
\end{center}
\vspace{-0.2cm}
\caption{Analysis of online template update mechanism.} 
\label{tab:ab2}
\vspace{-0.4cm}
\end{table}

\paragraph{Online template update.}
We investigate the impact of the online transformer mechanism for regression. The results of the investigation are shown in Table.~\ref{tab:ab2}. In our design, the online targets queue is composed of a static targets memory and an online updating targets memory. We can derive that the size of the static targets memory is of slight influence. Compared with only using a static target, maintaining an \textbf{online targets template queue} improves the AUC by 0.9\% and the Precision by 1.2\%. It demonstrates the effectiveness of the online regression mechanism. Furthermore, we observe that the performance drops a little if without the \textbf{confidence based updating strategy} since the online targets may be unreliable, which proves the effectiveness of the scheme.
Furthermore, we visualize the online targets queue in Figure.~\ref{fig:online_queue}. It can be derived
that high confidence scores generally correspond to high-quality
samples, which proves that the online samples selection
strategy is useful.

\begin{figure}[t]
\centering
\includegraphics[width=7cm]{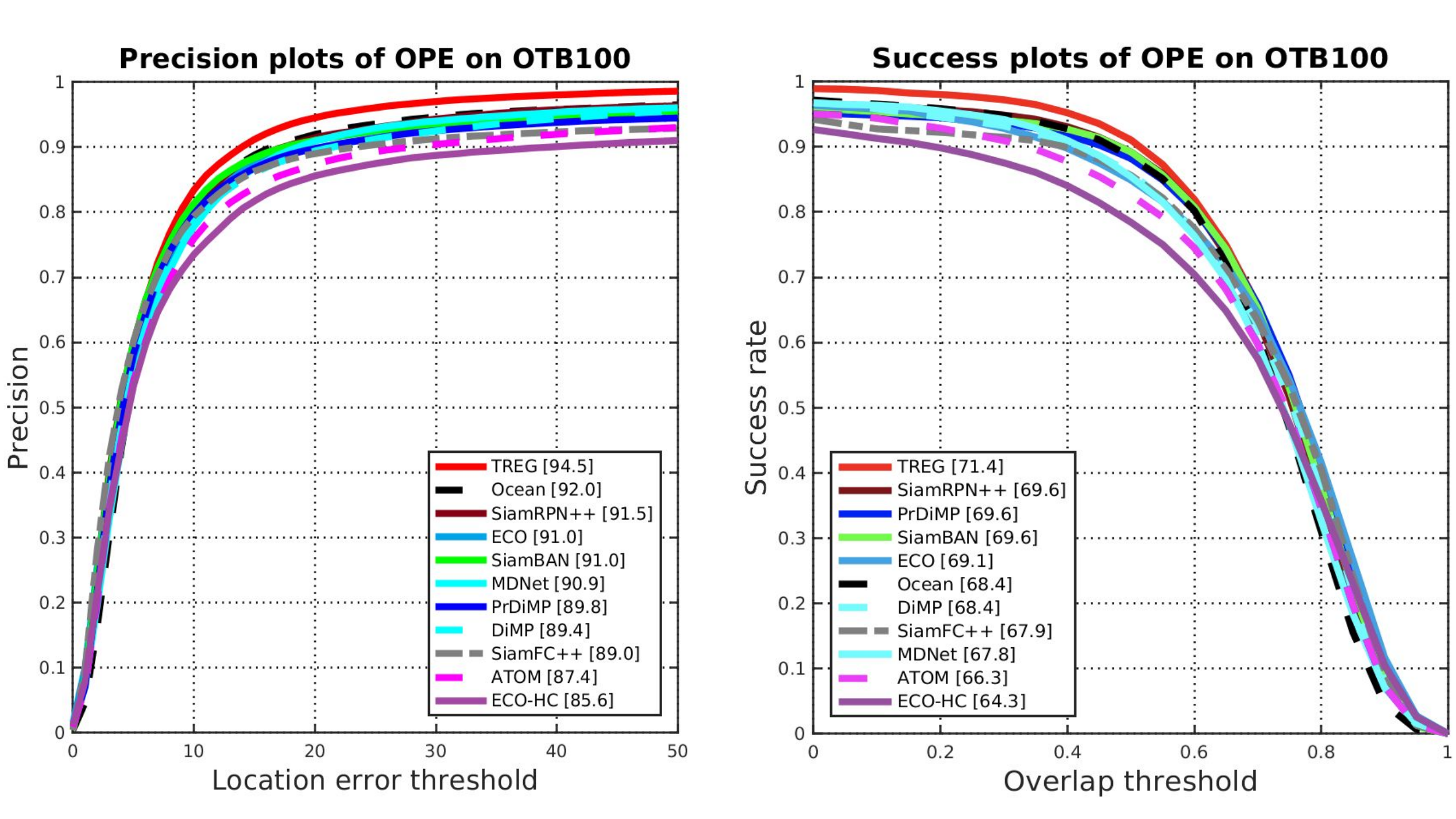}
\caption{Comparison results of trackers on OTB2015.
}
\label{fig:otb}
\vspace{-0.4cm}
\end{figure}

\begin{figure}[t]
\centering
\includegraphics[width=7cm]{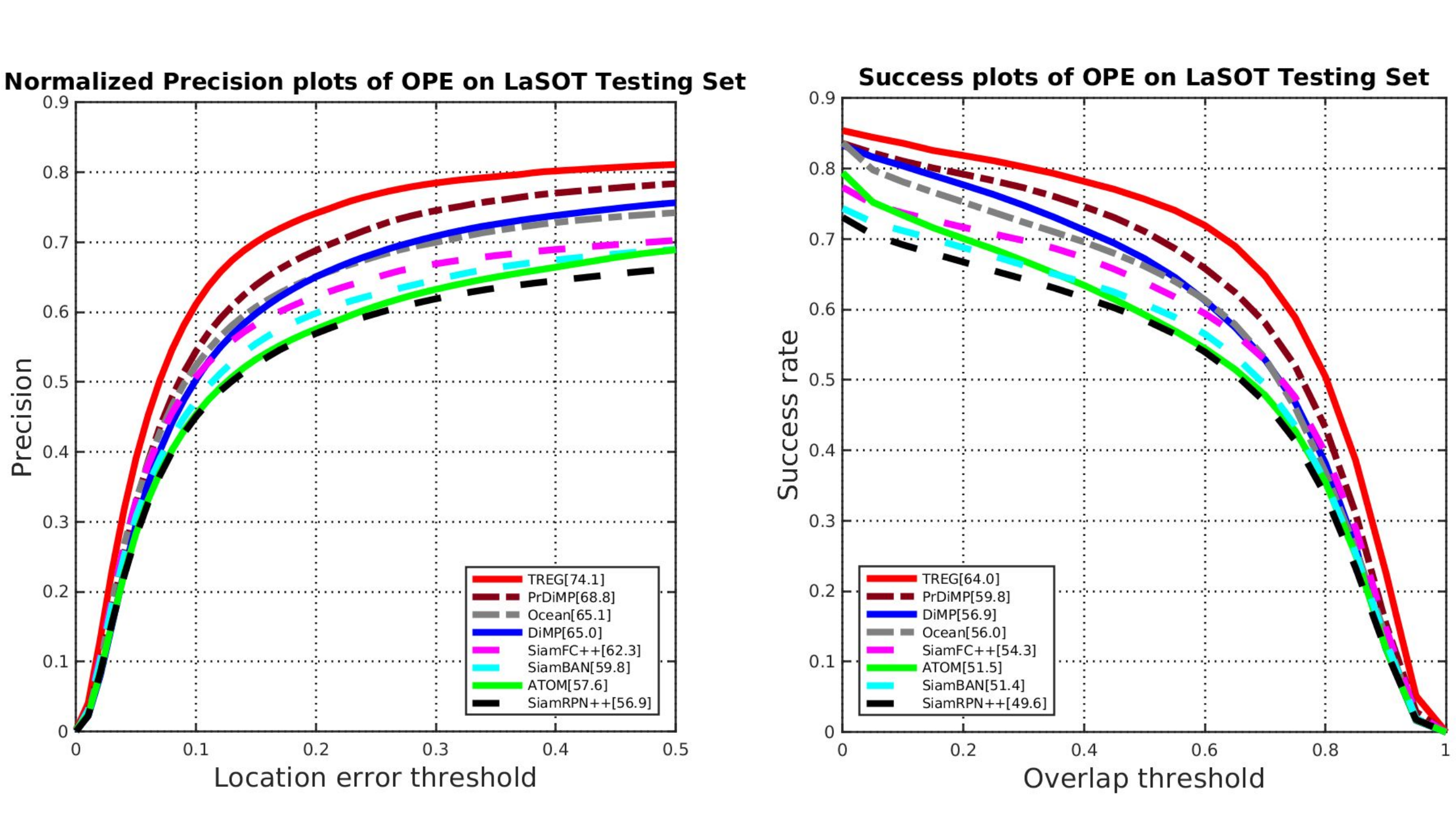}
\caption{Comparison results of trackers on LaSOT.}
\label{fig:lasot}
\vspace{-0.5cm}
\end{figure}

\subsection{Comparison with the state-of-the-art}
We test the proposed TREG on eight tracking benchmarks, including VOT2018~\cite{vot2018}, VOT2019~\cite{vot2019} LaSOT~\cite{lasot}, TrackingNet~\cite{trackingnet},UAV123~\cite{uav123}, GOT10k~\cite{got10k}, OTB100~\cite{otb} and NFS~\cite{nfs}, and compare our results with the state-of-the-art trackers.

\paragraph{OTB-100}
OTB100~\cite{otb} is a commonly used benchmark, which evaluates performance on Precision and AUC scores. Figure.~\ref{fig:otb} presents evaluation results of our TREG on both two metrics on OTB-100 benchmark. Achieving 71.4\% and 94.5\% in AUC score and precision score, our tracker reaches the state-of-the-art level w.r.t. other trackers in comparison.

\paragraph{LaSOT}

LaSOT~\cite{lasot} has 280 videos in its test set. We evaluate our TREG on the test set to validate its long-term capability. 
The Figure.~\ref{fig:lasot} shows that our TREG surpasses other real-time trackers with a large margin. Specifically, it achieves the top-ranked performance on AUC criteria of 64.0\% and Precision of 74.1\%. The remarkable performance suggests that our TREG not only adapts to long-term tracking, but keeps precision of online targets regression.

\paragraph{VOT2018}
Our TREG is tested on the VOT2018~\cite{vot2018} dataset consisting of 60 videos in comparison with the SOTA trackers.
As shown in Table.~\ref{tab:vot2018}, TREG achieves the performance on EAO criteria of 0.496 and Robustness of 0.098, which outperforms all state-of-the-art trackers. The improved Accuracy suggests that our TREG can generate precise bounding boxes.

\paragraph{VOT2019}

Our TREG is tested on the VOT2019~\cite{vot2019} dataset consisting of 60 videos in comparison with the SOTA trackers.
As shown in Table.~\ref{tab:vot2019}, TREG achieves the performance on EAO criteria of 0.391, Robustness of 0.221 and Accuracy of 0.603, which outperforms all state-of-the-art trackers. It suggests that our TREG can generate precise bounding boxes.

\begin{table}[t]
\begin{center}
\fontsize{7}{9}\selectfont  
\setlength{\tabcolsep}{1mm}{
\begin{tabular}{c c c c c c c c}
\toprule
                &ATOM&SiamRPN++&DiMP&PrDiMP&CGACD&Ocean&TREG\cr
                   &~\cite{atom}   &~\cite{siamrpnPlus}& ~\cite{dimp}&~\cite{prdimp}  &~\cite{CGACD}       &~\cite{ocean}   &\cr
\midrule
    Accuracy   &0.590&0.600&0.597&\textcolor{red}{\bf0.618}&\textcolor{blue}{\bf0.615}&0.592& 0.612\cr
    Robustness    &0.204&0.234&0.153&0.165&0.172&\textcolor{blue}{\bf0.117}&\textcolor{red}{\bf 0.098}\cr
    EAO    &0.401&0.414&0.440&0.442&0.449&\textcolor{blue}{\bf0.489}&\textcolor{red}{\bf 0.496}\cr
\bottomrule
\end{tabular}}
\end{center}
\vspace{-0.2cm}
\caption{Comparison with state-of-the-art trackers on VOT2018.} 
\label{tab:vot2018}
\vspace{-0.1cm}
\end{table}

\begin{table}[t]
\begin{center}
\fontsize{7}{9}\selectfont  
\setlength{\tabcolsep}{0.6mm}{
\begin{tabular}{c c c c c c c c}
\toprule
                &ATOM&SiamRPN++&DiMP50&SiamMask&SiamBAN&Ocean&TREG\cr
                   &~\cite{atom}   &~\cite{siamrpnPlus}& ~\cite{dimp}&~\cite{siammask} &~\cite{siamban}  &~\cite{ocean}   &\cr
\midrule
    Accuracy  &\textcolor{red}{\bf0.603} &0.599&0.594 &0.594 &\textcolor{blue}{\bf0.602} &0.594 & \textcolor{red}{\bf 0.603}\cr
    Robustness  &0.411  &0.482&\textcolor{blue}{\bf0.278} &0.461 &0.396 &0.316 &\textcolor{red}{\bf 0.221}\cr
    EAO   &0.292 &0.285 &\textcolor{blue}{\bf0.379} &0.287 &0.327 &0.350 &\textcolor{red}{\bf 0.391}\cr
\bottomrule
\end{tabular}}
\end{center}
\vspace{-0.2cm}
\caption{Comparison with state-of-the-art trackers on VOT2019.}
\label{tab:vot2019}
\vspace{-0.1cm}
\end{table}

\begin{table}[t]
\begin{center}
\fontsize{7}{9}\selectfont  
\setlength{\tabcolsep}{0.45mm}{
\begin{tabular}{cc c c c c c c c}
\toprule
                &ATOM&SiamRPN++&DiMP&PrDiMP&CGACD&SiamFC++&TREG\cr
                   &~\cite{atom}   &~\cite{siamrpnPlus}& ~\cite{dimp}&~\cite{prdimp}  &~\cite{CGACD}       &~\cite{siamfc++}   &\cr
\midrule
    Prec.(\%)    &64.8&69.4&68.7&70.4&69.3&\textcolor{blue}{\bf 70.5}&\textcolor{red}{\bf 75.0}\cr
    Norm. Prec.(\%)     &77.1&80.0&80.1&\textcolor{blue}{\bf 81.6}&80.0&80.0&\textcolor{red}{\bf 83.8}\cr
    Succ.(\%)   &70.3&73.3&74.0&\textcolor{blue}{\bf 75.8}&71.1&75.4&\textcolor{red}{\bf 78.5}\cr
\bottomrule %
\end{tabular} }%
\end{center}
\vspace{-0.2cm}
\caption{Comparison with state-of-the-art trackers on TrackingNet.} 
\label{tab:trackingnet}
\vspace{-0.4cm}
\end{table}

\paragraph{TrackingNet}

TrackingNet~\cite{trackingnet} provides over 30K videos with more than 14 million dense bounding box annotations. The videos are sampled from YouTube, covering target categories and scenes in real life. We validate TREG on its test set and achieve a remarkable improvement on all three metrics. Our TREG is proved to improve tracking performance on the large scale benchmark.

\begin{table}[pt]
\begin{center}
\fontsize{7}{9}\selectfont  
\setlength{\tabcolsep}{0.8mm}{
\begin{tabular}{cc c c c c c c}
\toprule
                &ATOM&DiMP&SiamFC++&SiamRCNN&PrDiMP&OCEAN&TREG\cr
                &~\cite{atom} &~\cite{dimp} &~\cite{siamfc++} &~\cite{siamrcnn} &~\cite{prdimp} &~\cite{ocean} &\cr
\midrule
    SR$_{0.50}$    &63.4&71.7&69.5&72.8&\textcolor{blue}{\bf 73.8}&72.1&\textcolor{red}{\bf 77.8}\cr
    
    SR$_{0.75}$     &40.2&49.2&47.9&\textcolor{red}{\bf 59.7}&54.3&-&\textcolor{blue}{\bf 57.2}\cr
    
    AO   &55.6&61.1&59.5&\textcolor{blue}{\bf 64.9}&63.4&61.1&\textcolor{red}{\bf 66.8}\cr
\bottomrule %
\end{tabular} }%
\end{center}
\vspace{-0.2cm}
\caption{Comparison with state-of-the-art trackers on GOT10k.} 
\label{tab:got10k}
\vspace{-0.1cm}
\end{table}

\begin{table}[t]
\begin{center}
\fontsize{7}{9}\selectfont  
\setlength{\tabcolsep}{0.55mm}{
\begin{tabular}{c c c  c c c c c}
\toprule
                &ATOM&SiamRPN++&SiamCAR&SiamBAN&DiMP&PrDiMP&TREG\cr
                   &~\cite{atom}  &~\cite{siamrpnPlus}& ~\cite{siamcar}&~\cite{siamban}  &~\cite{dimp}       &~\cite{prdimp}   &\cr
\midrule
    Precision(\%)   &-&80.7&76.0&\textcolor{blue}{\bf 83.3}&-&-&\textcolor{red}{\bf 88.4}\cr
    AUC(\%)     &64.4&61.3&61.4&63.1&65.4&\textcolor{red}{\bf 68.0}&\textcolor{blue}{\bf 66.9}\cr
\bottomrule
\end{tabular}}
\end{center}
\vspace{-0.2cm}
\caption{Comparison with state-of-the-art trackers on UAV123.}
\vspace{-0.1cm}
\label{tab:uav123}
\end{table}

\begin{table}[t]
\begin{center}
\fontsize{7}{9}\selectfont 
\setlength{\tabcolsep}{0.6mm}{
\begin{tabular}{c c c c c c c c c c c}
\toprule
                &CCOT&ECO&UPDT&ATOM&SiamBAN&DiMP&PrDiMP&KYS&TREG\cr
                &~\cite{ccot}      &~\cite{eco}   &~\cite{updt}    &~\cite{atom}    &~\cite{siamban}       &~\cite{dimp}    &~\cite{prdimp}      &~\cite{kys} &  \cr
\midrule
    AUC(\%)     &48.8&46.6&53.7&58.4&59.4&62.0&\textcolor{blue}{\bf63.5}&\textcolor{blue}{\bf 63.5} &\textcolor{red}{\bf 66.6}\cr
\bottomrule
\end{tabular}}
\end{center}
\vspace{-0.2cm}
\caption{Comparison with state-of-the-art trackers on NFS.} 
\label{tab:nfs}
\vspace{-0.3cm}
\end{table}

\paragraph{GOT10k}
GOT10k~\cite{got10k} is a large-scale dataset with over 10000 video segments and has 180 segments for the test set. Apart from generic classes of moving objects and motion patterns, the object classes in the train and test set are zero-overlapped. Our TREG obtain state-of-the-art performance on the test split.

\paragraph{UAV123}
UAV123~\cite{uav123} is a large dataset containing 123 Sequences with average sequence length of 915 frames, which is captured from low-altitude UAVs. Table.~\ref{tab:uav123} shows our results on UAV123 dataset.
Our proposed TREG outperforms the previous best reported result in precision. For AUC score, our TREG achieves 66.9\%, which is close to PrDiMP. 

\paragraph{NFS}
NFS dataset~\cite{nfs} contains a total of 380K frames in 100 videos from real world scenarios. We evaluate our TREG on the 30 FPS version of this dataset. As shown in Table.~\ref{tab:nfs}, Our TREG outperforms all previous approaches by a significant margin. Our result demonstrates the effectiveness of TREG for accurate regression.

\section{Conclusions}
We proposed a target-aware transformer, termed as TREG, to transfer an regression component from detection into tracking. 
We formulated the process of target integration for regression as a pixel-wise feature transformation guided by the target features. 
Additionally, we explored an efficient online regression mechanism which maintains an updating targets memory and selects the confident samples.
Our approach provided accurate target estimation while being robust against distractor objects in the scene, outperforming previous methods on eight datasets.

{\small
\bibliographystyle{ieee_fullname}
\bibliography{egbib}
}

\end{document}